\documentclass{ifacconf}

\usepackage{graphicx}      % include this line if your document contains figures
\usepackage{epsfig}
\graphicspath{ {./images/} }
\usepackage{natbib}        % required for bibliography
\usepackage{url}
\usepackage{cuted}
\usepackage{amsmath} % assumes amsmath package installed
\usepackage{amssymb}  % assumes amsmath package installed
\newcommand{\RomanNumeralCaps}[1]{\MakeUppercase{\romannumeral #1}}
\renewcommand{\thetable}{\Roman{table}}
\usepackage{booktabs}
\usepackage{subcaption}

\begin{document}
\begin{frontmatter}

\title{Fast Online Adaptive Neural MPC via Meta-Learning
\thanksref{footnoteinfo}} 

\thanks[footnoteinfo]{This research was supported by the National Science Foundation (CNS 2237577 and CMMI 1940950).}

\author[First]{Yu Mei,}
\author[First]{Xinyu Zhou,} 
\author[Second]{Shuyang Yu,} 
\author[First]{Vaibhav Srivastava,} 
\author[First]{and Xiaobo Tan}

\address[First]{Department of Electrical and Computer Engineering, \\
Michigan State University, East Lansing, MI 48824 USA \\
(e-mail: \{meiyu1, zhouxi63, vaibhav, xbtan\}@msu.edu)}

\address[Second]{Department of Computer Science and Engineering, \\
Michigan State University, East Lansing, MI 48824 USA \\
(e-mail: yushuyan@msu.edu)}

\begin{abstract}                % Abstract of not more than 250 words.
Data-driven model predictive control (MPC) has demonstrated significant potential for improving robot control performance in the presence of model uncertainties. However, existing approaches often require extensive offline data collection and computationally intensive training, limiting their ability to adapt online. To address these challenges, this paper presents a fast online adaptive MPC framework that leverages neural networks integrated with Model-Agnostic Meta-Learning (MAML). Our approach focuses on few-shot adaptation of residual dynamics—capturing the discrepancy between nominal and true system behavior—using minimal online data and gradient steps. By embedding these meta-learned residual models into a computationally efficient \textit{L4CasADi}-based MPC pipeline, the proposed method enables rapid model correction, enhances predictive accuracy, and improves real-time control performance. We validate the framework through simulation studies on a Van der Pol oscillator, a Cart-Pole system, and a 2D quadrotor. Results show significant gains in adaptation speed and prediction accuracy over both nominal MPC and nominal MPC augmented with a freshly initialized neural network, underscoring the effectiveness of our approach for real-time adaptive robot control.

\end{abstract}


\end{frontmatter}
%===============================================================================

\section{Introduction}

Data-driven model predictive control (MPC) has gained significant attention in recent years due to its ability to improve control performance in the presence of model uncertainties, unmodeled dynamics, and external disturbances. Various forms of data-driven MPC have been proposed, including learning-based MPC \citep{hewing2020learning, ren2022tutorial}, Gaussian Process MPC \citep{hewing2019cautious}, Koopman operator-based MPC \citep{korda2018linear}, Data-Enabled Predictive Control \citep{coulson2019data} and No-Regret MPC \citep{zhou2024koopman, zhou2024simultaneous}. These approaches leverage data to build predictive models that replace or augment traditional physics-based system dynamics within the MPC framework, enabling more accurate control in complex or partially known environments. Data-driven MPC has been successfully applied to a range of robotic systems, including quadrotors \citep{torrente2021data, salzmann2023real}, autonomous vehicles \citep{rosolia2018data}, and soft robots \citep{bruder2024koopman, wang2024mechanical}. 

While data-driven MPC methods have shown great potential, a major limitation lies in their reliance on extensive offline training, which often requires immense data collection and significant computational effort prior to deployment. This poses significant challenges for real-time adaptation in robotic systems operating under dynamic or uncertain conditions—particularly in safety-critical scenarios where data collection is limited or costly. Furthermore, solving optimization problems involving data-driven models can incur high computational cost, making real-time execution difficult on embedded hardware with constrainted processing capabilities. 

In this paper, we focus on learning-based MPC that leverages neural networks (NNs) to model system uncertainties, due to their demonstrated ability to accurately capture complex nonlinear ordinary differential equations (ODEs). This type of regression problems is also known as a Neural ODE \citep{chen2018neural} in the machine learning community. To enable seamless integration of NNs into the optimal control problem (OCP), the recent \textit{L4CasADi} framework \citep{salzmann2023real,salzmann2024learning} was introduced to convert PyTorch-trained NNs into symbolic expressions compatible with the CasADi optimization backend. This allows for differentiable and computationally efficient MPC formulations using learned dynamics, even for large and complex models. The framework has demonstrated the capability to execute real-time neural MPC and has been applied to various motion planning tasks \citep{jacquet2024n, gao2024integrated}. 

Despite the aforementioned progress, prior applications of the \textit{L4CasADi} framework have primarily focused on deploying fully trained models without adaptation during execution. These approaches still require substantial offline data collection and training. In contrast, in this paper we propose an online adaptive neural MPC method that enables real-time fine-tuning of the dynamics model during control execution, enabling the controller to handle model mismatch and disturbances without relying heavily on offline datasets. 

When a large neural network is fine-tuned online to adapt to an unknown system, standard techniques typically require substantial data and many gradient updates to achieve satisfactory performance. Within an MPC framework, this slow adaptation becomes problematic: although the physical system responds quickly, the model fails to keep pace because deep architectures generally require large amounts of data to learn effectively. As a result, the controller may rely on outdated or inaccurate dynamics, degrading overall performance. A more efficient strategy is therefore needed to enable rapid, few-shot adaptation suitable for real-time control.

Model-Agnostic Meta-Learning (MAML), introduced by \citet{finn2017model}, is a meta-learning algorithm that embodies the principle of ``learning to learn." The objective of meta-learning is to train a model across a distribution of tasks such that it can quickly adapt to new tasks using only a small number of training samples and gradient steps. Owing to its few-shot and fast adaptation capability, MAML is particularly well-suited for online adaptive control applications. As a result, it has begun to attract growing interest from the control~\citep{muthirayan2025meta, richards2023control} and robotics~\citep{tsuchiya2024online, tang2023meta} communities. Most of recent work focus on training meta-learned control policies directly, while relatively few explore how meta-learning can enhance model-based controllers. Recent efforts have integrated meta-learning with deep neural state-space models (NSSMs) to model unknown dynamical systems and perform optimal control~\citep{chakrabarty2023meta, yan2024mpc}. The most relevant work is by~\citet{sanghvi2024occam}, which employs Bayesian recursive estimation via meta-learning to learn prior predictive models that can quickly adapt to online data. However, its control performance may be sensitive to poor initialization due to the stochastic nature of the optimizer.

In this paper, we propose a fast online adaptive neural MPC framework for robotic systems based on Model-Agnostic Meta-Learning (MAML). Specifically, our approach learns residual dynamics, defined as the discrepancy between the nominal model and actual system behavior. By meta-learning neural network parameters, our method achieves rapid adaptation of residual models using minimal online data and few gradient updates. To validate the effectiveness of the proposed method, we first demonstrate online adaptation and model prediction performance using the Van der Pol oscillator. Additionally, we compare our MAML-MPC approach against two baseline MPC controllers on both a Cart-Pole system and a 2D quadrotor in a high-fidelity physics simulator. Compared to conventional fine-tuning techniques, our framework significantly accelerates adaptation, enabling accurate predictive modeling and prompt compensation for residual dynamics during real-time control.

The remainder of this paper is organized as follows. Section~\RomanNumeralCaps{2} introduces the problem statement and notation. Section~\RomanNumeralCaps{3} details the specific robotic system considered, describes the proposed online adaptive neural MPC framework, and outlines the MAML-based model preparation pipeline. Simulation results are presented in Section~\RomanNumeralCaps{4}. Finally, Section~\RomanNumeralCaps{5} provides conclusions and future work.

\section{Problem Statement}
We address the problem of real-time optimal control for nonlinear systems with partially known dynamics. Specifically, we consider a general continuous-time, nonlinear dynamical system described by:
\begin{subequations} \label{eq:1}
\begin{align}
\dot{x}(t) &= f(x(t), u(t)), \label{eq:1a} \\
 &= f_{\text{nom}}(x(t), u(t)) + f_{\text{res}}(x(t), u(t)), \label{eq:1b}
\end{align}
\end{subequations}
where $x(t) \in {\mathbb{R}^n}$ denotes the system state and $u(t) \in {\mathbb{R}^m}$ is the control input. And $f:{\mathbb{R}^n} \times {\mathbb{R}^m} \to {\mathbb{R}^n}$ represents the true (unknown) system dynamics. The function $f_{\text{nom}}:{\mathbb{R}^n} \times {\mathbb{R}^m} \to {\mathbb{R}^n}$ captures the known or approximate nominal dynamics, while $f_{\text{res}}:{\mathbb{R}^n} \times {\mathbb{R}^m} \to {\mathbb{R}^n}$ accounts for unknown residual dynamics, attributed to, for example, unmodeled dynamics or parameter errors. 

The goal is to design a control policy that achieves desired closed-loop performance, despite the unknown $f_{\text{res}}$. To this end, we propose an online learning-based MPC approach, where the residual dynamics $f_{\text{res}}$ is learned in real-time and combined with the nominal model to improve prediction accuracy.

 Using the direct multiple shooting method \citep{bock1984multiple}, the MPC problem is formulated as a nonlinear program. At each time step $t$, the controller solves the following finite-horizon optimal control problem over a prediction horizon $t_f$:
\begin{align} \label{eq:costfunction}
\min_{{\mathbf{u}}_{0:N - 1}} \quad & \sum_{k=0}^{N-1} \ell(x_k, u_k) + m(x_N) \notag \\
\text{s.t.} \quad & x_{k+1} = \phi (x_k, u_k, f, \delta t), \quad \forall k = 0, \dots, N-1, \notag \\
& x_0 = x(t) \notag \\
& g(x_k, u_k) \leq 0, \quad \forall k = 0, \dots, N-1
\end{align}         
Here, $\ell(x_k, u_k)$ is the stage cost, $m(x_N)$ is the terminal cost, and $\phi(\cdot)$ is a general numerical integrator that discretizes the continuous-time dynamics $f$ over time steps $\delta t = t_f / N$. A common example of a numerical integrator is the fourth-order Runge--Kutta method. $N$ denotes the number of discretized steps in the prediction horizon. The constraint $g(x_k, u_k)$ models general nonlinear state and input constraints. The optimization is initialized at the current system state $x(t)$ and produces an optimal control sequence $\mathbf{u}_{0:N-1}^*$.  
Only the first control input $u_0$ is applied to the system at time $t$, following the receding horizon principle.

To ensure closed-loop adaptation, the MPC must operate in real time, with the system dynamics $f$ continuously refined as the residual model $f_{\text{res}}$ is learned online.

\section{Methods}
\subsection{Robotic System Setup}
For common mechanical or robotic systems, the general dynamics and MPC cost function defined in the previous section can be instantiated using system-specific state definitions and task objectives. Without loss of generality, let $x = [{x_1}, {x_2}]^T$ denote the system state, where $x_1 \in \mathbb{R}^{n_x}$ represents the position and $x_2 \in \mathbb{R}^{n_x}$ the velocity. The full state vector $x \in \mathbb{R}^{n}$ thus satisfies $n = 2n_x$. We assume that both components can be directly measured from the onboard sensors. Therefore, the derivative of the state $\dot x = {[{\dot x_1},{\dot x_2}]^T} = {[{x_2},{\dot x_2}]^T}$. The nominal model approximates this derivative as $f_{\text{nom}}(x, u) = [x_2, \hat{\dot{x}}_2]^T$, where $\hat{\dot{x}}_2$ denotes the estimated acceleration from the nominal dynamics. Since the velocity can be measured directly, the residual model only needs to compensate the acceleration term. The residual dynamics can be constructed as ${f_\text{res}}(x,u) = {[0,{f_{{\text{NN}}}}(x,u;\theta )]^T}$, where ${f_{{\text{NN}}}}(x,u;\theta):{\mathbb{R}^n} \times {\mathbb{R}^m} \to {\mathbb{R}^{{n_x}}}$ represents the learning-based nonlinear function via a neural network (NN) with parameters $\theta$. As a result, one can rewrite Eq.~\eqref{eq:1b} as:
\begin{align} \label{eq:3}
\dot{x}(t) &= f_{\text{nom}}(x(t), u(t)) + f_{\text{res}}(x(t), u(t)) \notag \\
          &= \begin{bmatrix}
              x_2 \\
              \hat{\dot{x}}_2
            \end{bmatrix}
           + \begin{bmatrix}
              0 \\
              f_{\text{NN}}(x, u; \theta)
            \end{bmatrix} \notag \\
          &= \begin{bmatrix}
              x_2 \\
              \hat{\dot{x}}_2 + f_{\text{NN}}(x, u; \theta)
            \end{bmatrix} 
\end{align}

In the MPC problem setup, we use a common quadratic cost function suitable for both stabilization and tracking tasks. The stage cost can be specified as
\begin{equation} \label{eq:4}
    \ell(x_k, u_k) = \left\| x_k - x_{\text{ref},k} \right\|_Q^2 + \left\| u_k - u_{\text{ref},k} \right\|_R^2,
\end{equation}
and the terminal cost is defined as
\begin{equation} \label{eq:5}
m(x_N) = \left\| x_N - x_{\text{ref},N} \right\|_Q^2,
\end{equation}
where $\| z \|_M^2 := z^\top M z$ denotes the squared weighted norm with matrix $M$, and $Q \geqslant 0$, $R \geqslant 0$ are the state and input weighting matrices. The terms $x_{\text{ref},k}$ and $u_{\text{ref},k}$ denote the reference state and control input at time step $k$, respectively. To ensure actuator limitations are respected, the control input is constrained by box bounds:
\begin{equation} \label{eq:6}
u_{\min}^{(i)} \leq u_k^{(i)} \leq u_{\max}^{(i)}, \quad \forall k = 0, \dots, N-1,\; \forall i = 1, \dots, n_u,
\end{equation}
where \(u_k^{(i)}\) denotes the \(i\)-th control input at time step \(k\), and \(u_{\min}^{(i)}\), \(u_{\max}^{(i)}\) define the lower and upper bounds for each input dimension \(i\).

\subsection{Online Adaptive Neural MPC}
The proposed online adaptive neural MPC framework is an indirect adaptive controller that learns and updates the unknown residual dynamics model in real time using on-the-fly collected data. 

\begin{figure}[h]
    \centering
    \includegraphics[scale=0.58]{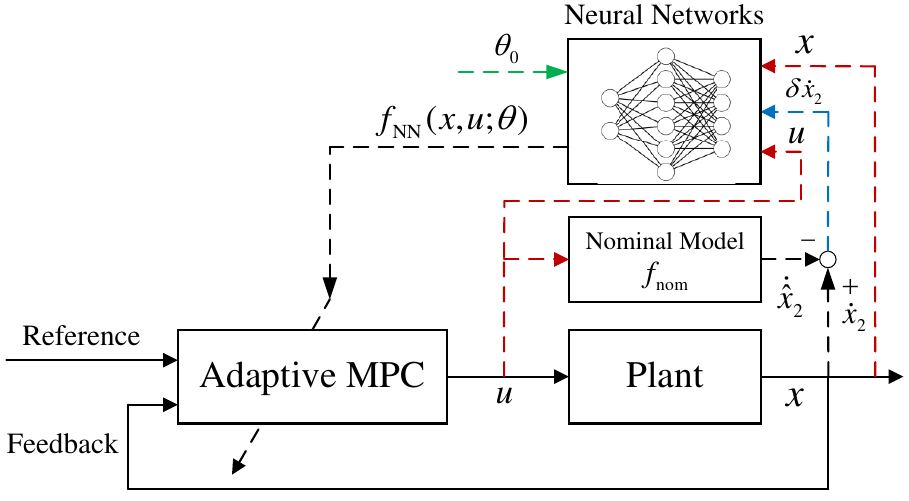}
    \caption{Block diagram of the proposed online adaptive neural MPC framework.}
    \label{fig:adaptive_mpc_diagram}
\end{figure}

As illustrated in Fig~\ref{fig:adaptive_mpc_diagram}, the acceleration compensation term $f_{\text{NN}}(x, u; \theta)$ is learned using a supervised learning approach. The NN takes the current state $x$ and control input $u$ as inputs (indicated by the red dashed lines), and the training label is the observed discrepancy $\delta \dot{x}_2$ between the nominal acceleration prediction $\hat{\dot{x}}_2$ and the actual system acceleration $\dot{x}_2$, which is shown as a blue dashed line. The true acceleration $\dot{x}_2$ is estimated by differentiating the measured velocity. The loss function used for training the NN is the mean absolute error (MAE) of the observed discrepancy $\delta \dot{x}_2$. The residual model $f_{\text{NN}}(x, u; \theta)$ aims to minimize the observed discrepancy. Simultaneously, the adaptive MPC controller updates the additive dynamics model and computes the optimal control input based on the refined model. To reduce computational overhead, the residual model $f_{\text{NN}}(x, u; \theta)$ is trained and updated using mini-batches of data rather than continuously. In other words, the NN is updated periodically at fixed time intervals ${T_{{\text{update}}}}$.

In this work, the NN is built and trained using \textit{PyTorch}, which offers automatic gradient computation through its dynamic computation graph (\textit{autograd}). The MPC problem is formulated symbolically using \textit{CasADi} and solved efficiently with the \textit{acados} solver, both of which are widely used for nonlinear optimal control and algorithmic differentiation. To enable the integration of learned dynamics into the MPC framework, we leverage the \textit{L4CasADi} interface to bridge \textit{PyTorch} and \textit{CasADi}. The PyTorch model is trained and updated online, and \textit{L4CasADi} is programmed to regenerate the corresponding \textit{CasADi} computation graph for the adaptive NN.

\begin{figure*}[t]
    \centering
    \includegraphics[scale=0.52]{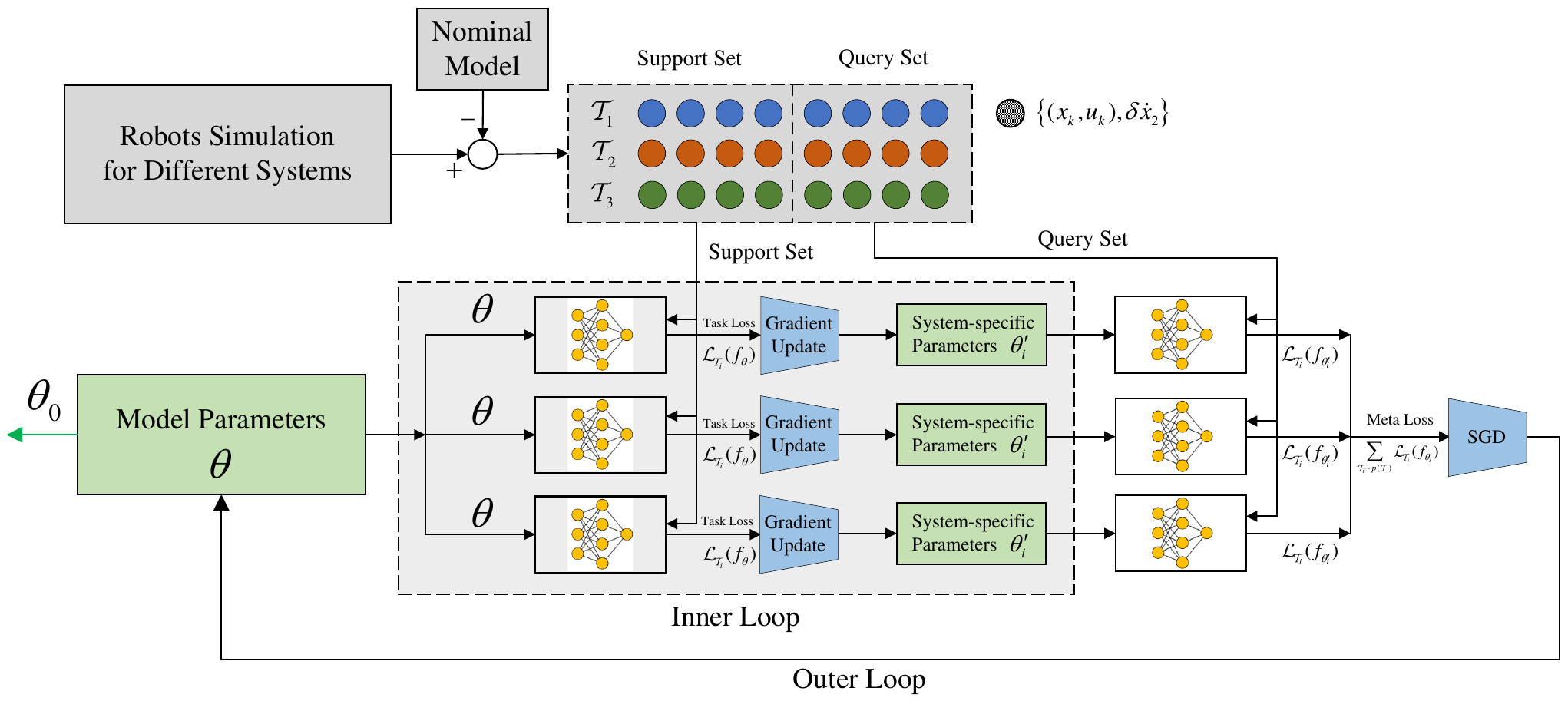}
    \caption{Overview of the MAML-based meta-learning approach for robotic systems, illustrating the inner-loop adaptation of task-specific parameters \(\theta'_i\) and the outer-loop optimization of shared meta-parameters \(\theta\). This framework enables rapid adaptation to new system dynamics using limited data.}
    \label{fig:MetaLearning}
\end{figure*}

The proposed online adaptive neural MPC framework is agnostic to the specific NN architecture, as long as the network is implemented in \textit{PyTorch} and remains traceable and differentiable. While different architectures may yield varying performance depending on the system dynamics and task complexity, architecture construction is not the focus of this study. To keep the framework simple and efficient, a multilayer perceptron (MLP) is adopted as the residual model $f_{\text{NN}}(x, u; \theta)$ throughout this work.

\subsection{Fast Adaptation via Model-Agnostic Meta-Learning}

Fast adaptation is critical for maintaining responsive and reliable control in robotic applications. However, training the proposed neural network online for residual dynamics can be time-consuming, especially when using deep or large NN architectures for complex systems. To facilitate fast adaptation, MAML is employed to pretrain the NN parameters. This meta-learning strategy enables the model to efficiently adapt to previously unseen system dynamics—termed ``tasks" in the original MAML formulation—using only a small number of online samples, a capability commonly referred to as few-shot learning. 

Figure~\ref{fig:MetaLearning} illustrates the workflow for pretraining a neural network model to capture residual dynamics with rapid adaptation capability using a MAML-based meta-learning approach. To align with the MAML framework, we formulate the regression of residual dynamics under varying physical properties or operating conditions as a set of distinct tasks. Specifically, a collection of robotic systems is simulated, where each task \(\mathcal{T}_i\) corresponds to a system exhibiting different dynamics (e.g., variations in mass, friction, or actuator latency). By comparing with the nominal model prediction, one can obtain a tuple $\left\{ {({x_k},{u_k}),\delta {{\dot x}_2}} \right\}$, represented as colorful dots in the upper part of Fig.~\ref{fig:MetaLearning}. Notably, in the $K$-shot learning setting, only $2K$ samples are required for training each task ${\mathcal{T}_i}$ in the meta-learning phase. Among them, $K$ samples is used as support set $\mathcal{D}_i^{\text{support}}$ for inner-loop adaption and another $K$ samples is used as query set $\mathcal{D}_i^{\text{query}}$ for outer-loop meta-update. In the inner loop, the model parameters $\theta$ are adapted to a specific task $\mathcal{T}_i$ by performing single gradient descent steps on the support data $\mathcal{D}_i^{\text{support}}$, yielding system-specific parameters 
$\theta_i'$:
\begin{equation}
    \theta_i' = \theta - \alpha \nabla_\theta \mathcal{L}_{\mathcal{T}_i}^{\text{support}}(f_\theta)
\end{equation}
where $\alpha$ is the inner-loop learning rate, and $\mathcal{L}_{\mathcal{T}_i}^{\text{support}}$ is the loss computed on the support data $\mathcal{D}_i^{\text{support}}$ for task $\mathcal{T}_i$.
In the outer loop, the meta-objective is to minimize the query loss after inner-loop adaptation. The meta-parameters $\theta$ are updated by aggregating gradients across tasks via stochastic gradient descent (SGD):
\begin{equation}
\theta \leftarrow \theta - \beta \nabla_\theta \sum_{\mathcal{T}_i \sim p(\mathcal{T})} \mathcal{L}_{\mathcal{T}_i}^{\text{query}}(f_{\theta_i'})
\end{equation}
where $\beta$ is the meta learning rate, and $\mathcal{L}_{\mathcal{T}_i}^{\text{query}}$ is the loss evaluated on the query set $\mathcal{D}_i^{\text{query}}$ using the adapted parameters $\theta_i'$.

After completing the meta-training process, the meta-learned initial parameters \(\theta_0\) are well-initialized for fast online adaptation. This initialization enables the neural network to be fine-tuned efficiently using a small number of real-time samples, making it particularly suitable for the online adaptive neural MPC framework described in Section 3.2.

\section{Simulation Examples and Results}
In order to validate the proposed framework, four simulation examples are used to evaluate its performance in terms of model prediction accuracy, tracking performance, and adaptation speed. The simulations cover a range of dynamical systems with increasing complexity, including the Van der Pol oscillator, Cart-Pole system, 2D quadrotor stabilization, and 2D quadrotor trajectory tracking. To benchmark the effectiveness of our method, three MPC controllers are compared: (i) a nominal MPC using only the nominal model, (ii) a nominal MPC augmented with a newly initialized MLP for residual dynamics compensation, and (iii) a nominal MPC augmented with a meta-learned MLP (MetaMLP) for fast adaptation, which is the proposed method. Sample code for constructing the meta-learned residual dynamics and improving the performance MPC in real time can be found at the Github repository. \footnote{Code: \url{https://github.com/yu-mei/MetaResidual-MPC.git}} A demonstration video is also available. \footnote{Video: \url{https://youtu.be/4K2QeBxWcWA}}

\subsection{Van der Pol Oscillator}
To evaluate whether the proposed method can accelerate the learning of residual dynamics, we consider the well-known Van der Pol oscillator with an uncertain damping coefficient. The system dynamics are governed by the following ordinary differential equation (ODE):
\begin{equation}
\begin{bmatrix}
\dot{x}_1(t) \\
\dot{x}_2(t)
\end{bmatrix}
=
\begin{bmatrix}
x_2(t) \\
\mu \left(1 - x_1(t)^2\right)x_2(t) - x_1(t)
\end{bmatrix},
\end{equation}
where $\mu  \geqslant 0$ is the damping ratio of the oscillator. In our setup, the true system dynamics correspond to $\mu_{\text{real}} = 0.2$, while the nominal model assumes $\mu_{\text{nom}} = 0.7$.

To best simulate the real-time controller scenario, we use the models from three MPC controllers to predict the trajectory \([x_1, x_2]\) over a 1-second prediction horizon. The residual dynamics models (MLP and MetaMLP) are fine-tuned online using observed error from the previous 1-second window, and the observation considers the measurement Gaussian noise $\mathcal{N}(0,\, 0.025^2)$. Given the system operates at 50~Hz, each fine-tuning step uses a batch of 50 samples. Consequently, the MetaMLP is updated with \(K = 50\) samples per adaptation. Other architecture and hyperparameter settings can be found in Table~\RomanNumeralCaps{1}. In the meta-training phase, we collected tasks with varying \(\mu\) values, sampled in increments of 0.1 from 0 to 1. For each task, we simulated the oscillator dynamics starting from a random initial condition $x_0 \in [-2, 2]^2$ over a duration of 10 seconds. The initial model parameters \(\theta_0\) were trained using the meta-learning approach described in Section~3.3, with the hyperparameters listed in Table~\RomanNumeralCaps{1}(b).

\begin{table}[h] 
    \centering
    \caption{Architecture and hyperparameter settings for residual dynamics models and the meta-learning phase.}
    \vspace{0.5em}
    \begin{subtable}[t]{0.48\textwidth}
        \centering
        \caption{Architecture and online fine-tuning parameters for both MLP and MetaMLP}
        \begin{tabular}{p{5.3cm} p{2cm}}
            \toprule
            \textbf{Parameter} & \textbf{Value} \\
            \midrule
            MLP architecture           & [2, 64, 64, 1] \\
            Prediction horizon \(t_f\) & 1 s \\
            Fine-tune period \(T_\text{update}\) & 1 s \\
            Fine-tune epochs           & 50 \\
            Fine-tune batch size \(K\) & 50 \\
            \bottomrule
        \end{tabular}
    \end{subtable}
    \hfill
    \vspace{5pt}
    \begin{subtable}[t]{0.48\textwidth}
        \centering
        \caption{Offline meta-training hyperparameters}
        \begin{tabular}{p{5.3cm} p{2cm}}
            \toprule
            \textbf{Parameter} & \textbf{Value} \\
            \midrule
            Inner learning rate \(\alpha\) & \(1 \times 10^{-2}\) \\
            Meta learning rate \(\beta\)   & \(1 \times 10^{-3}\) \\
            Training epochs                & 20000 \\
            Training sample size \(K\)     & 50 \\
            \bottomrule
        \end{tabular}
    \end{subtable}
\end{table}

We compare three models on a system with \(\mu_{\text{real}} = 0.2\), starting from the initial condition \(x_0 = [0.5, 0.5]^\top\). After averaging over 10 trials with different random initializations of the neural networks and noise realizations, the root mean square error (RMSE) between the predicted and true trajectories is \(0.0963 \pm 0.0772\) for the nominal model, \(0.0495 \pm 0.0245\) for the nominal model augmented with a newly initialized MLP (Nominal + Residual MLP), and \(0.0377 \pm 0.0144\) for the nominal model augmented with a meta-learned MLP (Nominal + Meta-Residual MLP). We also visualize the time evolution of the states \(x_1(t)\) and \(x_2(t)\), as well as the phase portrait (\(x_1\) vs. \(x_2\)) with predicted trajectories from a representative trial, as shown in Fig.~3. It can be observed that the nominal model fails to accurately predict the future trajectory. The nominal model augmented with residual MLP improves prediction quality by mitigating residual dynamics through online learning. However, the nominal model augmented with residual MetaMLP achieves the most accurate and consistent predictions, demonstrating both fast adaptation and high fidelity compared to the other two baseline methods.

The prediction speed is also evaluated: the nominal model achieves a prediction frequency of at least 10,500\,Hz, while both the nominal model augmented with a residual MLP and the model augmented with a MetaMLP achieve prediction frequencies of at least 6,200\,Hz. All evaluations were performed on a laptop equipped with an AMD 9900X processor and 32\,GB of RAM.
\begin{figure}[h]
    \centering
    \includegraphics[width=0.85\columnwidth]{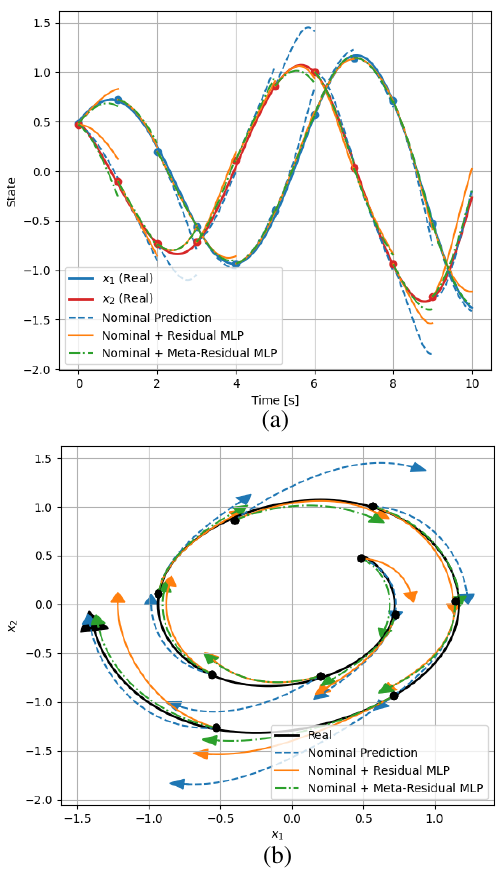}
    \caption{Prediction comparison using three models in MPC controllers. 
    (a) Time evolution of states \(x_1(t)\) and \(x_2(t)\). 
    (b) Phase portrait (\(x_1\) vs. \(x_2\)) illustrating trajectory predictions.}
    \label{fig:VanderPol_results}
\end{figure}

\subsection{Cart-Pole System}
\begin{figure}[h]
    \centering
    \includegraphics[width=0.9\columnwidth]{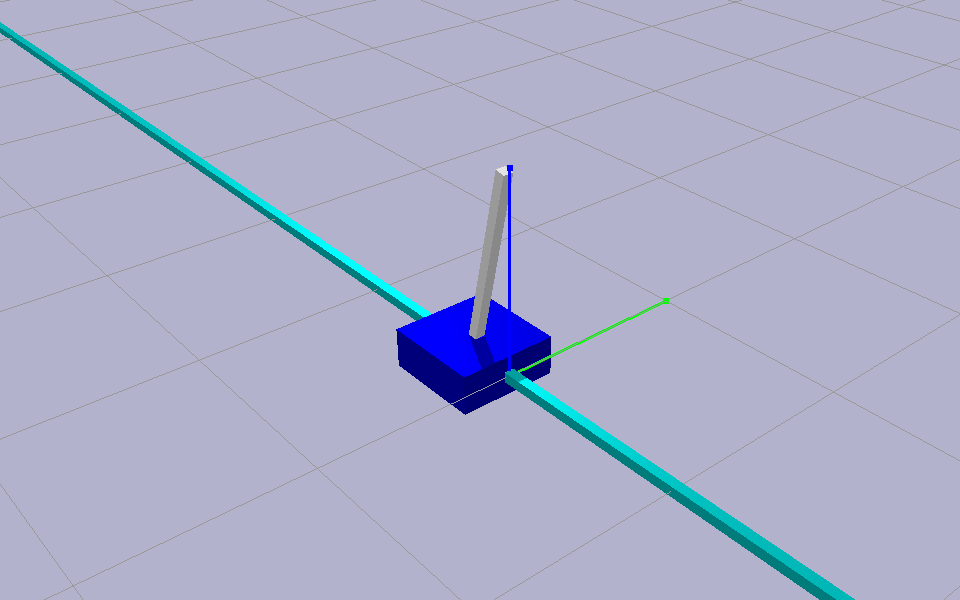}
    \caption{The snapshot of Cart-Pole system in PyBullet}
    \label{fig:Cartpole}
\end{figure}
After validating the model prediction accuracy, here we apply the online adaptive neural MPC for stabilizing a Cart-Pole system, as shown in Fig.~\ref{fig:Cartpole}. In this system, the state vector is $x = {[p,\dot p,\theta ,\dot \theta ]^T}$, where $p$ is the horizontal position of the cart, $\theta$ is the angle of the pole with respect to the vertical direction. In frictionless case, the dynamics can be formulated as:
\begin{equation}
\begin{aligned}
  \ddot{p} &= \frac{F + m_p l (\dot{\theta}^2 \sin \theta - \ddot{\theta} \cos \theta)}{m_c + m_p}, \\
  \ddot{\theta} &= \frac{g \sin \theta + \cos \theta \left( \frac{-F - m_p l \dot{\theta}^2 \sin \theta}{m_c + m_p} \right)}{l \left( \frac{4}{3} - \frac{m_p \cos^2 \theta}{m_c + m_p} \right)},
\end{aligned}
\end{equation}
where $m_c$ and $m_p$ are the mass of cart and pole respectively, $l$ is the pole length and $g$ is the gravity acceleration. $F$ is the input force for moving cart in the horizontal direction.

In this case, the true system parameters are ${m_c} = 1,{m_p} = 0.1,l = 0.5$, while we assume the nominal system parameters are scaled to 66\% of the true system. We apply three MPC controllers with control frequency 50 Hz for stabilizing the Cart-Pole to $x_\text{ref}=[0,0,0,0]^T$, with shared MPC parameters listed in the Table~\RomanNumeralCaps{2}(b). The NN architecture and hyperparameters are reported in the Table~\RomanNumeralCaps{2}(a). In the meta-training phase, we collect data from up to 50 tasks, each defined by a different \((m, I_{yy})\) pair, where the parameters vary within the range \([0.75, 2.0]\) relative to their nominal values. Each task is simulated with 3 different initial positions over a duration of 10--15 seconds. The simulations are conducted using the \textit{safe-control-gym} framework~\citep{yuan2022safe} within the \textit{PyBullet} physics simulator. The training and deployment is same as Cart-Pole system.

\begin{table}[h]
    \centering
    \caption{Architecture and parameter settings for the Cart-Pole system.}
    \vspace{0.5em}

    \begin{subtable}[t]{0.48\textwidth}
        \centering
        \caption{MLP/MetaMLP architecture and meta-training parameters}
        \begin{tabular}{p{2.2cm} p{1.6cm} | p{2.2cm} p{1.3cm}}
            \toprule
            \multicolumn{2}{c|}{\textbf{MLP + Online Fine-Tuning}} & \multicolumn{2}{c}{\textbf{Offline Meta-Training}} \\
            \midrule
            \textbf{Parameter} & \textbf{Value} & \textbf{Parameter} & \textbf{Value} \\
            \midrule
            MLP arch. & {[5,64,64,64,2]} & Inner LR \(\alpha\) & \(1 \times 10^{-3}\) \\
            \(T_\text{update}\) & 0.2 s & Meta LR \(\beta\) & \(1 \times 10^{-4}\) \\
            Fine-tune epochs & 20 & Training epochs & 20000 \\
            Batch size \(K\) & 20 & Sample size \(K\) & 20 \\
            \bottomrule
        \end{tabular}
    \end{subtable}
    \label{tab:controller_config}

    \vspace{1em}

    \begin{subtable}[t]{0.48\textwidth}
        \centering
        \caption{MPC parameter setup}
        \begin{tabular}{p{5.3cm} p{2.5cm}}
            \toprule
            \textbf{Parameter} & \textbf{Value} \\
            \midrule
            Prediction horizon time \(t_f\)  & 1 \\
            Prediction horizon steps \(N\)   & 20 \\
            Weighting matrix \(Q\)           & diag[5, 0.1, 5, 0.1] \\
            Weighting matrix \(R\)           & 0.1 \\
            \bottomrule
        \end{tabular}
    \end{subtable}
    \label{tab:controller_config}
\end{table}
To evaluate performance, we conducted 100 repeated experiments with randomized initial states. In terms of stabilization success rate, the nominal MPC failed to stabilize the system in all trials (0\% success rate). The nominal MPC augmented with a residual MLP achieved a success rate of 98\%, while the MetaMLP-based controller successfully stabilized the pole in all 100 trials (100\% success rate). In addition to reliability, stabilization speed was also assessed. The MetaMLP-based controller required \(2.55 \pm 0.49\) seconds on average to stabilize the pole, whereas the residual MLP-based controller took \(3.72 \pm 1.36\) seconds. 

These results demonstrate that the MetaMLP-based online adaptive MPC achieves both robust stabilization and significantly faster adaptation. The high success rate and reduced stabilization time indicate that the meta-learned initialization enables the neural network to quickly capture the task-specific residual dynamics using only a few online samples. This highlights the benefit of incorporating meta-learning for real-time control in systems with variable or uncertain dynamics.

\subsection{2D Quadrotor System}
\begin{figure}[h]
    \centering
    \includegraphics[width=0.9\columnwidth]{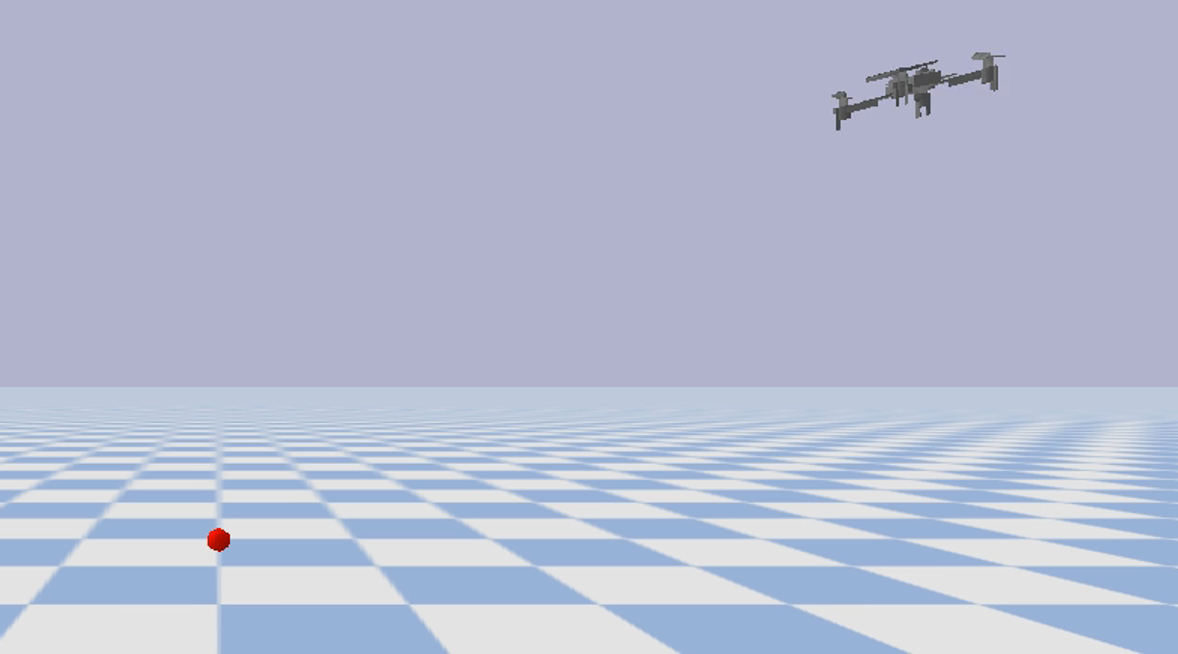}
    \caption{The snapshot of 2D quadrotor system in PyBullet}
    \label{fig:2D_quadrotor}
\end{figure}
The 2D quadrotor is a quadrotor whose motion is restricted within a vertical plane, as shown in Fig.~\ref{fig:2D_quadrotor}. In this system, the state vector is defined as ${\mathbf{x}} = {[x,\dot x,z,\dot z,\theta ,\dot \theta ]^T}$, where $x$ and $z$ are the horizontal and vertical position coordinates of the quadrotor, respectively, and $\theta$ is the orientation angle relative to the horizontal axis.

We evaluate the proposed controller on stabilization and tracking tasks to determine its effectiveness on more complex and practical robotic systems. According to \cite{yuan2022safe}, the dynamics of the 2D quadrotor can be expressed as:

\begin{equation} \label{eq:quadrotor_dynamics}
\begin{aligned}
    \ddot{x} &= \frac{\sin\theta \,(T_1 + T_2)}{m}, \\
    \ddot{z} &= \frac{\cos\theta \,(T_1 + T_2)}{m} - g, \\
    \ddot{\theta} &= \frac{(T_2 - T_1) \, d}{I_{yy}},
\end{aligned}
\end{equation}
where $m$ is the mass of the quadrotor, $I_{yy}$ is the moment of inertia about the out-of-plane axis, and $g$ is gravitational acceleration. $T_1$ and $T_2$ denote the input thrust forces from the left and right sides of the drone, respectively, and $d$ stands for the distance between the two thrust forces.

In our experiments, the true system parameters are $m = 0.027$~kg and $I_{yy}=1.4\times10^{-5} \ \text{kg} \cdot \text{m}^2$, while the nominal mass is scaled to $66\%$ of the true mass and the $I_{yy}$ is scaled to $80\%$ of the ture value. The three MPC controllers are all implemented with the same MPC configuration used in Cart-Pole system as listed in the Table~\RomanNumeralCaps{2}(b). The NN hyperparameters are also the same as those used for Cart-Pole system, shown in Table~\RomanNumeralCaps{2}(a). The only difference is that the MLP architecture was changed to $[8,64,64,64,3]$ to match the input/output dimensions of the 2D quadrotor system. 

In the meta-training phase, we collect data from up to 100
task with varying values of $m$ and $I_{yy}$. The variation range for both parameters is \([0.75, 2.0]\) relative to their nominal values. After the meta-training, the meta-learned MLP is fed into the online adaptive MPC framework, with finetune interval $T_{\text{update}}=0.2$ s.

The three controllers are tested in the stabilization task and the tracking task of the 2D quadrotor system, which both demonstrate the fast online adaptation capability of the proposed Neural MPC framework via Meta-Learning.

\begin{figure}[h]
    \centering
    \includegraphics[width=0.95\columnwidth]{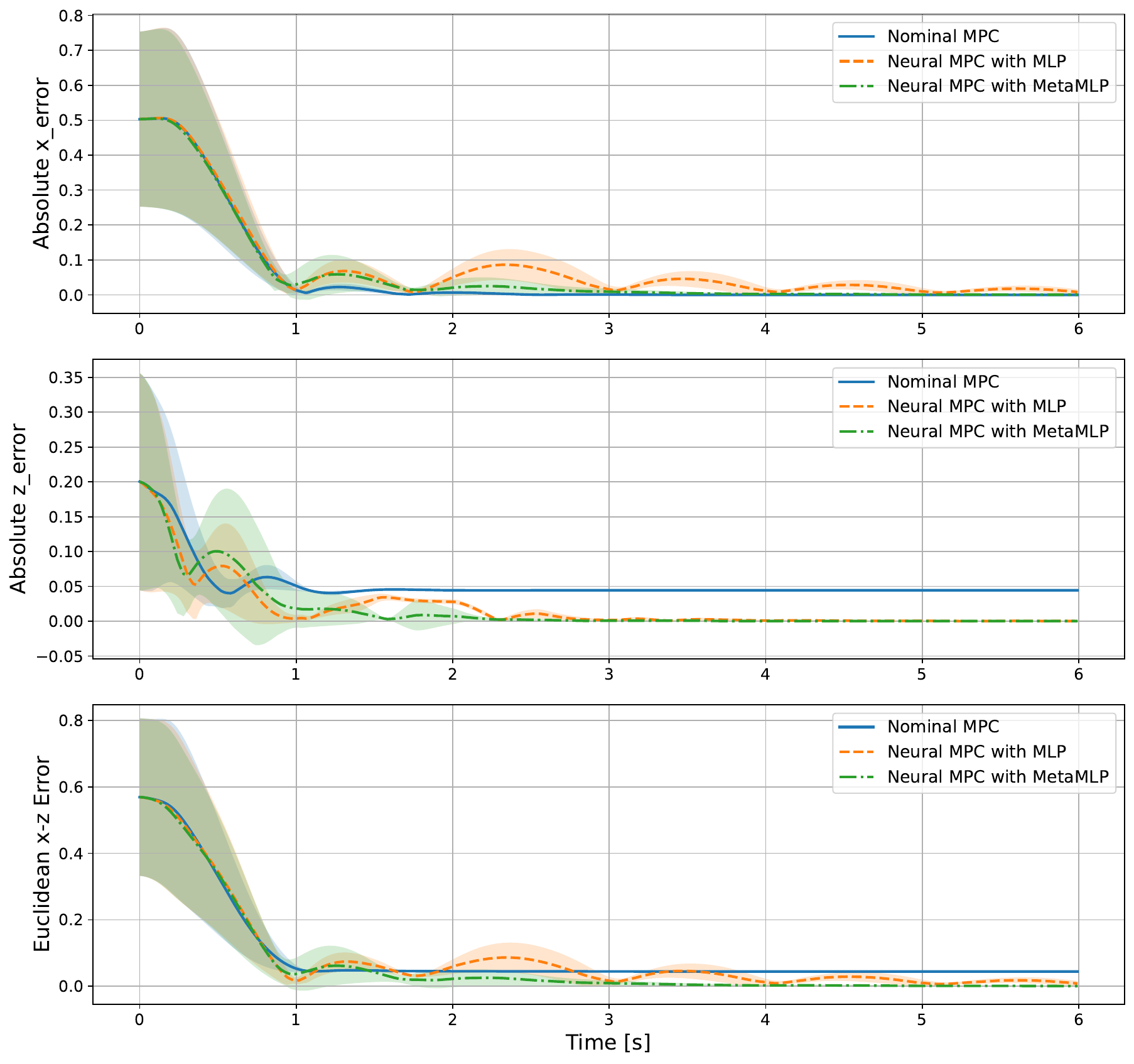}
    \caption{The comparison results of stabilizing the 2D quadrotor. The shaded regions represent standard deviations over 20 trials. (a) Averaged absolute errors of $x$ over time. (b) Averaged absolute errors of $z$ over time. (c) Averaged absolute Euclidean errors of $x-z$ over time.}
    \label{fig:2D_Stabilization}
\end{figure}

\textbf{Stabilization Task} For the stabilization task, the reference state is held constant at ${{\mathbf{x}}_{{\text{ref}}}} = {[0,0,1,0,0,0]^T}$. To evaluate the performance of the controllers, 20 repeated simulations with randomized initial conditions are conducted, the averaged errors are shown in Fig.~\ref{fig:2D_Stabilization}.
As illustrated in Fig.~\ref{fig:2D_Stabilization}, the nominal MPC is able to regulate the position rapidly; however, it exhibits a steady-state error in stabilizing \(z\) due to inaccuracies in the nominal model. The yellow trajectory represents the neural MPC with an uninitialized MLP, which achieves stabilization without steady-state error but shows increased oscillation. This behavior is attributed to the fact that the MLP must explore the state space and gradually learn the residual dynamics during execution. As shown in Fig.~\ref{fig:2D_Stabilization}(c), the Neural MPC with the meta-learned MLP completes the regulation task within 4 seconds, whereas the variant with the uninitialized MLP requires more than 6 seconds. Our proposed Neural MPC with the meta-learned MLP not only accelerates the adaptation process but also eliminates noticeable oscillations, owing to its ability to rapidly incorporate residual dynamics through few-shot learning.

\textbf{Tracking Task} Similar to the stabilization task, 20 repeated simulations are conducted to validate the controller's performance in tracking a circular reference trajectory. The circular path is centered at \((0, 1)\) with a radius of 0.5\,m and a period of 15\,s. The corresponding reference state is defined as:
\begin{equation}
\mathbf{x}_{\text{ref}}(t) =
\begin{bmatrix}
x_{\text{ref}}(t) \\
\dot{x}_{\text{ref}}(t) \\
z_{\text{ref}}(t) \\
\dot{z}_{\text{ref}}(t) \\
\theta_{\text{ref}}(t) \\
\dot{\theta}_{\text{ref}}(t)
\end{bmatrix}
=
\begin{bmatrix}
0.5 \cos\!\left(\frac{2\pi}{15}\, t\right) \\
-\frac{\pi}{15}\, \sin\!\left(\frac{2\pi}{15}\, t\right) \\
1 + 0.5 \sin\!\left(\frac{2\pi}{15}\, t\right) \\
\frac{\pi}{15}\, \cos\!\left(\frac{2\pi}{15}\, t\right) \\
0 \\
0
\end{bmatrix}.
\end{equation}

\begin{figure}[h]
    \centering
    \includegraphics[width=0.95\columnwidth]{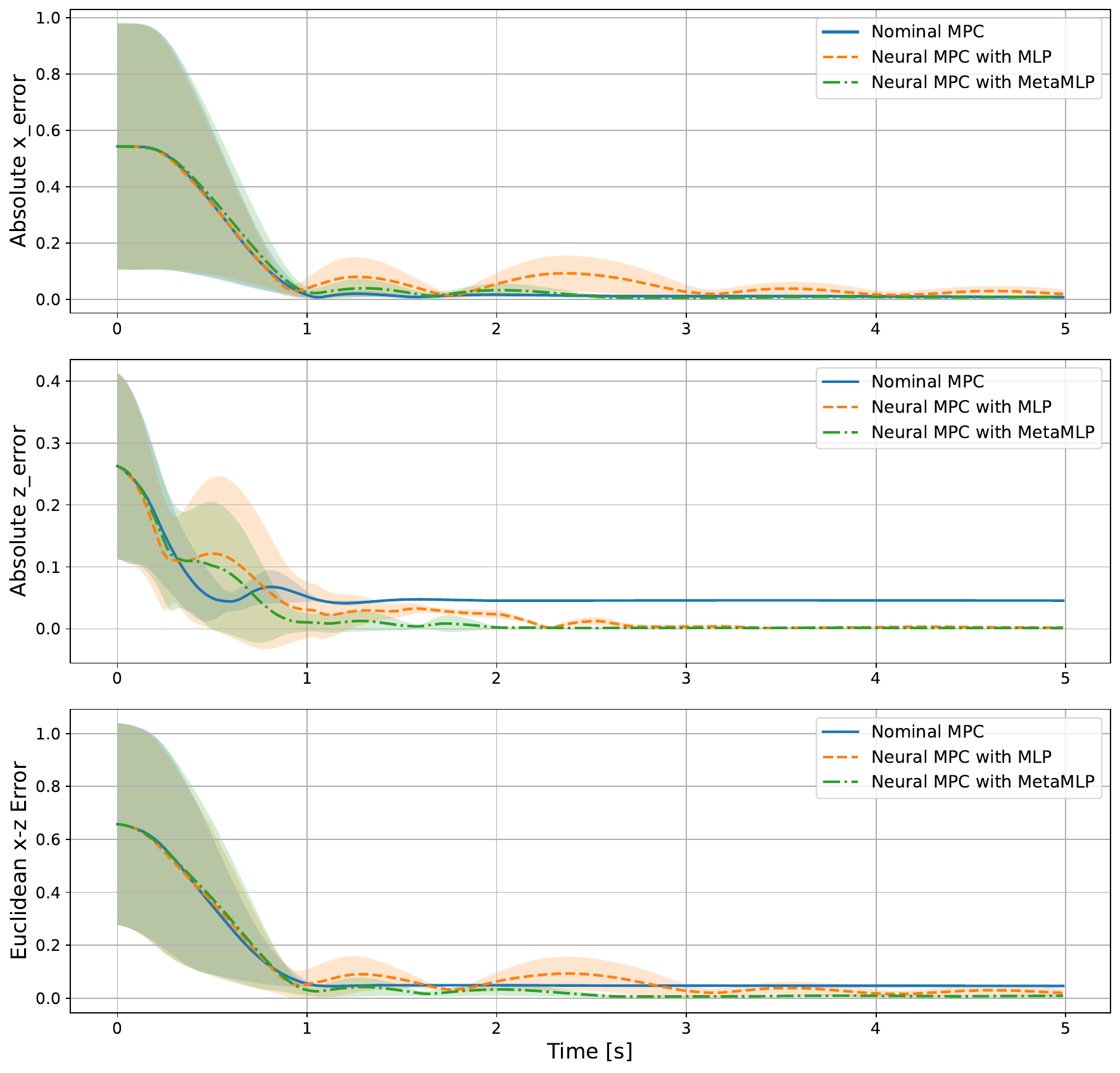}
    \caption{The comparison results of tracking circular reference. The shaded regions represent standard deviations over 20 trials. (a) Averaged absolute errors of $x$ over time. (b) Averaged absolute errors of $z$ over time. (c) Averaged absolute Euclidean errors of $x-z$ over time.}
    \label{fig:2D_Tracking}
\end{figure}

\begin{figure*}[t]
    \centering
    \includegraphics[scale=0.5]{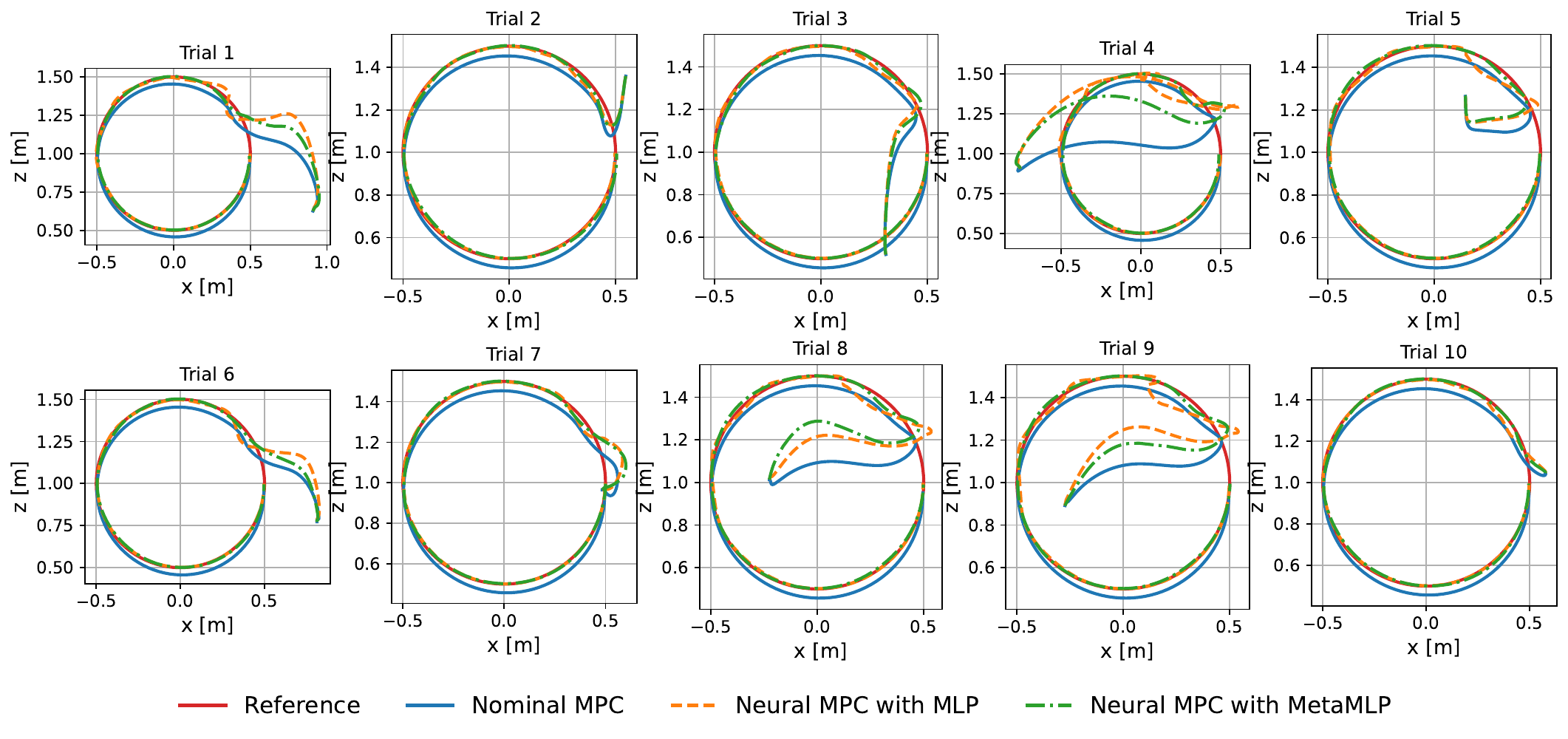}
    \caption{Trajectories of the 2D quadrotor tracking the circular reference in the \(x\)-\(z\) plane. Ten sample trials are visualized for each method.}
    \label{fig:2D_Tracking_trials}
\end{figure*}

The averaged errors are shown in Fig.~\ref{fig:2D_Tracking} and 10 trials of trajectories are visualized for comparison in Fig.~\ref{fig:2D_Tracking_trials}.
A similar trend is observed in the 2D trajectory tracking task, where the nominal MPC suffers from steady-state errors, and the Neural MPC with an uninitialized MLP exhibits larger oscillations. In contrast, the meta-learned MLP achieves fast and accurate tracking with minimal deviation from the reference trajectory.

In terms of computational efficiency, both the Cart-Pole and 2D quadrotor systems demonstrate that Neural MPC with either an MLP or a MetaMLP can sustain a control frequency of 45--50\,Hz. Notably, the online adaptation step serves as the computational bottleneck, running at approximately 2\,Hz, while all other components of the control pipeline operate at over 250\,Hz.

\section{Conclusion and Future Work}
This paper presents a fast online adaptive neural MPC framework for robotic systems using Model-Agnostic Meta-Learning (MAML) to enable efficient few-shot learning of residual dynamics. Integrated into the L4CasADi-based MPC pipeline, the approach achieves rapid adaptation and improved control performance. Simulation results on the Van der Pol oscillator, Cart-Pole, and 2D quadrotor show that our method outperforms nominal MPC and conventional neural MPC under model mismatch and uncertainty. A primary limitation of the proposed method lies in the computational bottleneck during the fine-tuning step, as the model update within a single control cycle can be time-consuming. This issue could be mitigated through multithreading or by optimizing the training frequency and number of epochs per update.

Future work will extend this framework to real-world robotic platforms to validate its hardware feasibility. In addition, we plan to investigate different neural network like RNN and LSTM for better time series data prediction.

\bibliography{ifacconf}             % bib file to produce the bibliography
                                                     % with bibtex (preferred)
                                                   
%\begin{thebibliography}{xx}  % you can also add the bibliography by hand

%\bibitem[Able(1956)]{Abl:56}
%B.C. Able.
%\newblock Nucleic acid content of microscope.
%\newblock \emph{Nature}, 135:\penalty0 7--9, 1956.

%\bibitem[Able et~al.(1954)Able, Tagg, and Rush]{AbTaRu:54}
%B.C. Able, R.A. Tagg, and M.~Rush.
%\newblock Enzyme-catalyzed cellular transanimations.
%\newblock In A.F. Round, editor, \emph{Advances in Enzymology}, volume~2, pages
%  125--247. Academic Press, New York, 3rd edition, 1954.

%\bibitem[Keohane(1958)]{Keo:58}
%\newblock \emph{Power and Interdependence: World Politics in Transitions}.
%\newblock Little, Brown \& Co., Boston, 1958.

%\bibitem[Powers(1985)]{Pow:85}
%\newblock Is there a way out?
%\newblock \emph{Harpers}, pages 35--47, June 1985.

%\bibitem[Soukhanov(1992)]{Heritage:92}
%\newblock \emph{{The American Heritage. Dictionary of the American Language}}.
%\newblock Houghton Mifflin Company, 1992.

%\end{thebibliography}
\end{document}